\pdfoutput=1

\documentclass[11pt]{article}

\usepackage[final]{acl}

\usepackage{times}
\usepackage{latexsym}

\usepackage[T1]{fontenc}

\usepackage[utf8]{inputenc}

\usepackage{microtype}

\usepackage{inconsolata}

\usepackage{graphicx}

\usepackage{booktabs}
\usepackage{amsmath}
\usepackage{algorithm}
\usepackage{algpseudocode}

\usepackage{multirow}
\usepackage{amssymb}
\usepackage{tabularx}
\usepackage{xcolor}
\usepackage{arydshln}
\usepackage{enumitem}
\usepackage{soul}
\usepackage{subcaption}

\title{CBP-Tuning: Efficient Local Customization for Black-box Large Language Models}

\author{
    Jiaxuan Zhao$^{1,2}$\footnotemark[1], 
    Naibin Gu$^{1,2}$\thanks{$\quad$ Equal Contribution.}, 
    Yuchen Feng$^{1,2}$, 
    Xiyu Liu$^{1,2}$, \\
    \textbf{Peng Fu}$^{1,2}$\thanks{$\quad$ Corresponding Author.}\textbf{,}  
    \textbf{Zheng Lin}$^{1,2}$\textbf{,} 
    \textbf{Weiping Wang}$^{1}$\\
    $^1$Institute of Information Engineering, Chinese Academy of Sciences, Beijing, China\\
    $^2$School of Cyber Security, University of Chinese Academy of Sciences, Beijing, China\\
   \texttt{\{zhaojiaxuan,gunaibin,fupeng\}@iie.ac.cn}
}

\begin{document}
\maketitle
\begin{abstract}
The high costs of customizing large language models (LLMs) fundamentally limit their adaptability to user-specific needs. Consequently, LLMs are increasingly offered as cloud-based services, a paradigm that introduces critical limitations: providers struggle to support personalized customization at scale, while users face privacy risks when exposing sensitive data. To address this dual challenge, we propose Customized Black-box Prompt Tuning (CBP-Tuning), a novel framework that facilitates efficient local customization while preserving bidirectional privacy. Specifically, we design a two-stage framework: (1) a prompt generator trained on the server-side to capture domain-specific and task-agnostic capabilities, and (2) user-side gradient-free optimization that tailors soft prompts for individual tasks. This approach eliminates the need for users to access model weights or upload private data, requiring only a single customized vector per task while achieving effective adaptation. Furthermore, the evaluation of CBP-Tuning in the commonsense reasoning, medical and financial domain settings demonstrates superior performance compared to baselines, showcasing its advantages in task-agnostic processing and privacy preservation.
\end{abstract}

\begin{figure}[ht]
  \centering
  \includegraphics[width=0.48\textwidth]{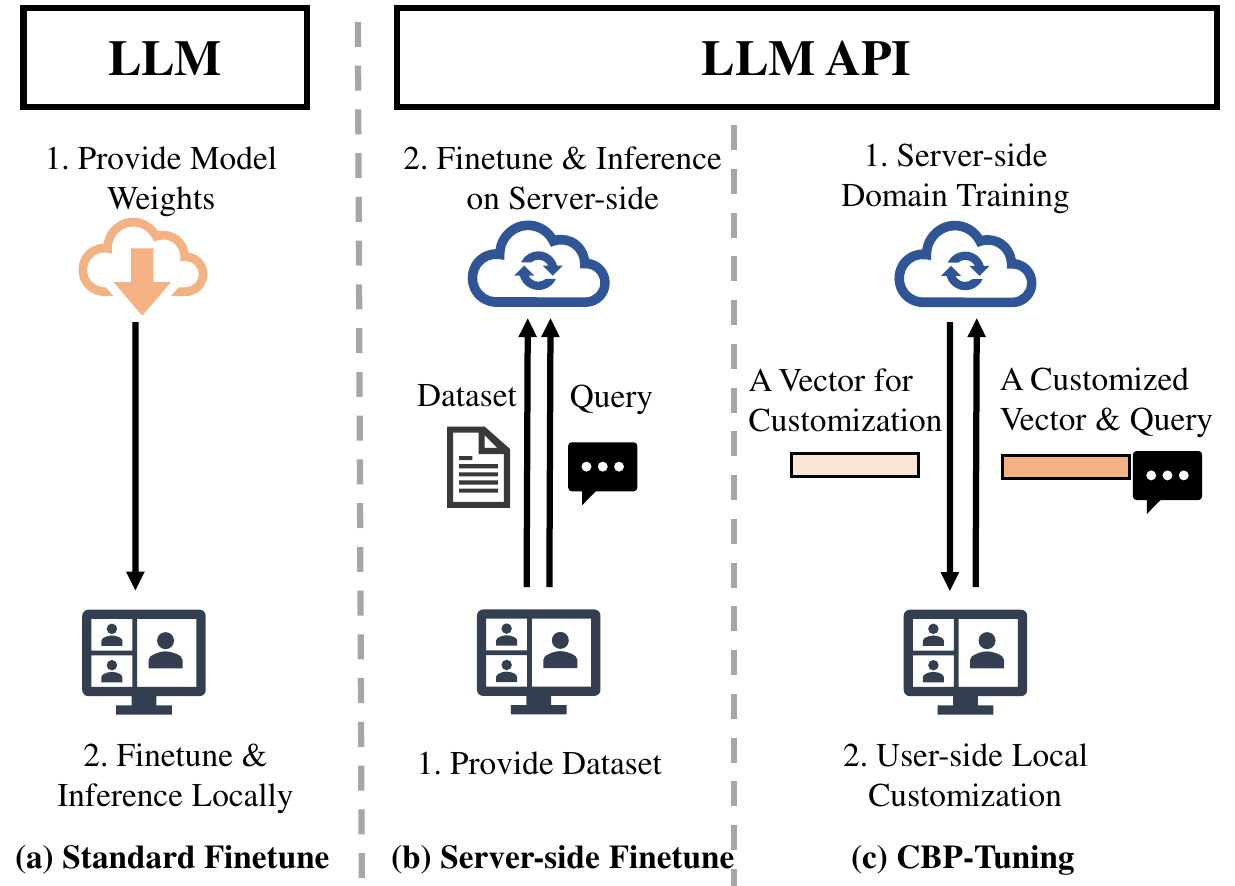}
  \caption{Schematic illustration of the comparison among three fine-tuning paradigms. (a) Left: Standard finetune transmits the model weights to the users.
  (b) Middle: Server-side finetune requires the users to provide the dataset for fine-tuning the model on the server side. 
  (c) Right: Our proposed method, CBP-Tuning, conducts domain training on the server side and allows users to customize a vector locally without transferring data or accessing the model weights. 
  }
  \label{fig:intro_pic}
\end{figure}

\section{Introduction}

Large language models (LLMs) have demonstrated extremely powerful performance in a wide range of tasks \cite{gpt,touvron2023llama,yang2024qwen2}. However, as these models become larger, the resources required for training and deployment become increasingly expensive~\cite{gu-etal-2024-light,feng2025divemoediversityenhancedreconstruction}, making it no longer feasible to fine-tune and deploy a separate model for each downstream task. This limits the ability of users to customize the model according to their specific needs. Parameter-efficient fine-tuning (PEFT) provides a promising solution by fine-tuning very few parameters while keeping most of the model parameters unchanged \cite{peft-survey,DBLP:journals/corr/abs-2407-05417,gu2025adaptoncethriveupdates, yang2025orthogonalfinetuningdirectpreference}. The PEFT technique includes adaption-based methods \cite{houlsby2019adapter,DBLP:journals/corr/abs-2110-04366}, re-parameterization-based methods \cite{hu2021lora,DBLP:conf/acl/Gu0000WS00025}, and prompt-based methods \cite{lester2021prompt-tuning,li2021prefix}, which enables a base model to serve multiple users simultaneously.

Despite the efficiency and effectiveness of methods such as Adapter \cite{houlsby2019adapter} and LoRA \cite{hu2021lora}, these approaches necessitate the server to uphold multiple PEFT modules for downstream tasks. For each batch inference task, a subset of these modules must be selected and assembled \cite{wen2023batched}, which is inconvenient in the multi-user scenario. In contrast, Prompt Tuning \cite{lester2021prompt-tuning} is a simpler approach that only requires a certain length of soft prompts to be prepended at the input layer. Before batch processing, users can provide the learned soft prompts and inputs for a specific task to the server. 

On the other hand, besides the cost of computation, LLMs also face the issue of privacy protection. Specifically, the server does not want users to access the full model weights, and users do not want to expose their data to the server. Most LLMs are released as services, and users can only access them through a black-box API. Black-Box Tuning \cite{sun2022black-box} describes this scenario as LMasS (Language Model as a Service), allowing users to optimize prompts using local black-box optimization methods. Despite their success, Black-box Tuning methods demonstrate limited versatility across tasks and LLMs \cite{zheng2024bpt-subspace}. In Offsite Tuning \cite{xiao2023offsite}, users can fine-tune the adapter for downstream data using an emulator sent from the server. In this setting, however, users still need to train and save the adapter parameters for each downstream task and upload them to the server for inference. This presents a challenge: 

\textbf{\textit{How to enable users to adapt the model without further retraining at the user side, while avoiding the need for server-side processing of task-specific parameters?}}

We aim to delegate the computationally intensive challenge of domain learning to server-side training, allowing users to focus only on optimizing customized tasks, thereby achieving efficient local customization. Therefore, we introduce a lightweight and flexible framework, CBP-Tuning (\textbf{C}ustomized \textbf{B}lack-box \textbf{P}rompt Tuning). As illustrated in Figure \ref{fig:intro_pic}, the server-side conducts domain training and sends a vector for customization to users, where the user can then use a black-box optimization method to adapt this vector to a specific task. Specifically, the server-side trains a prompt generator which is a feed-forward layer with a bottleneck architecture, receiving both instance-specific and task-specific input. With a domain-specific prompt generator in place, users can then fine-tune it in a black-box manner. Following the setting of Black-box Tuning \cite{sun2022black-box}, we project high-dimensional task-specific vectors into a smaller subspace. The user-side employs a black-box method to optimize the low-dimensional vector locally, thereby obtaining soft prompts tailored specifically for a given task. The user only needs to locally store a low-dimensional vector corresponding to each downstream task, without returning any additional parameters to the server-side.

Our contributions are summarized as follows: 

\begin{itemize}
    \item We propose a customized and lightweight fine-tuning framework called CBP-Tuning. It is a black-box optimization method based on prompt generators that achieves a balance between server-side and user-side, enabling efficient local customization.
    \item We design a prompt generator that takes the sum of task-specific vectors and instance-specific vectors as the input to the generator, combining both information.
    \item We conduct experiments using LLaMA-2-7B, Qwen-2.5-3B and LLaMA-2-13B to evaluate our method on general commonsense reasoning and domain-specific medical and financial tasks. The experimental results demonstrate that our approach outperforms baselines while offering lower costs and ensuring privacy.

\end{itemize}

\section{Preliminary: Black-box Tuning}

Black-Box Tuning (BBT)~\cite{sun2022black-box} can optimize the continuous prompt prepended to the input via derivative-free optimization. Considering a black-box LLM that predicts a batch of inputs $\mathbf{X}$ and outputs $\mathbf{Y}$, prompt tuning involves training continuous prompts $\mathbf{p}\in\mathbb{R}^{l*d}$ to achieve better performance when the model is fed the optimal prompt vector $\mathbf{p}^*$ together with the input, where $l$ is the length of the continuous prompt and $d$ is the model dimension. The objective of prompt tuning can be formulated as: 
\begin{align}
    \mathbf{p^*} = arg \min_{\mathbf{p}\in\mathcal{P}} \mathcal{L}(f(\mathbf{p}; \mathbf{X}), \mathbf{Y}),
\end{align}
where $f(\cdot)$ is the black-box LLM inference API, $\mathcal{L}(\cdot)$ is the loss function and $\mathcal{P}$ is a search space of prompts. In standard cases where a model can be accessed, the prompt $\mathbf{p}$ is optimized by gradient-based back-propagation. 

BBT leverages the low intrinsic dimensionality of LLMs by optimizing in a reduced subspace $\mathbf{z}\in\mathbb{R}^r\ (r\ll d)$ via random projection $\mathbf{A}\in\mathbb{R}^{d\times r}$:

\begin{align}
    \mathbf{z^*} = \arg\min_{\mathbf{z}} \mathcal{L}(f(\mathbf{Az}; \mathbf{X}), \mathbf{Y})
\end{align}

The Covariance Matrix Adaptation Evolution Strategy (CMA-ES) \cite{hansen2003cmaes} maintains a multivariate normal distribution $\mathcal{N}(\mathbf{m}^{(t)}, \sigma^{(t)}\mathbf{C}^{(t)})$ to sample candidate solutions $\mathbf{z}_i^{(t+1)}$. Through iterative population evaluation, the distribution parameters $(\mathbf{m}, \sigma, \mathbf{C})$ evolve toward high-performance regions.

\begin{figure}[tb]
  \centering
  \includegraphics[width=\linewidth]{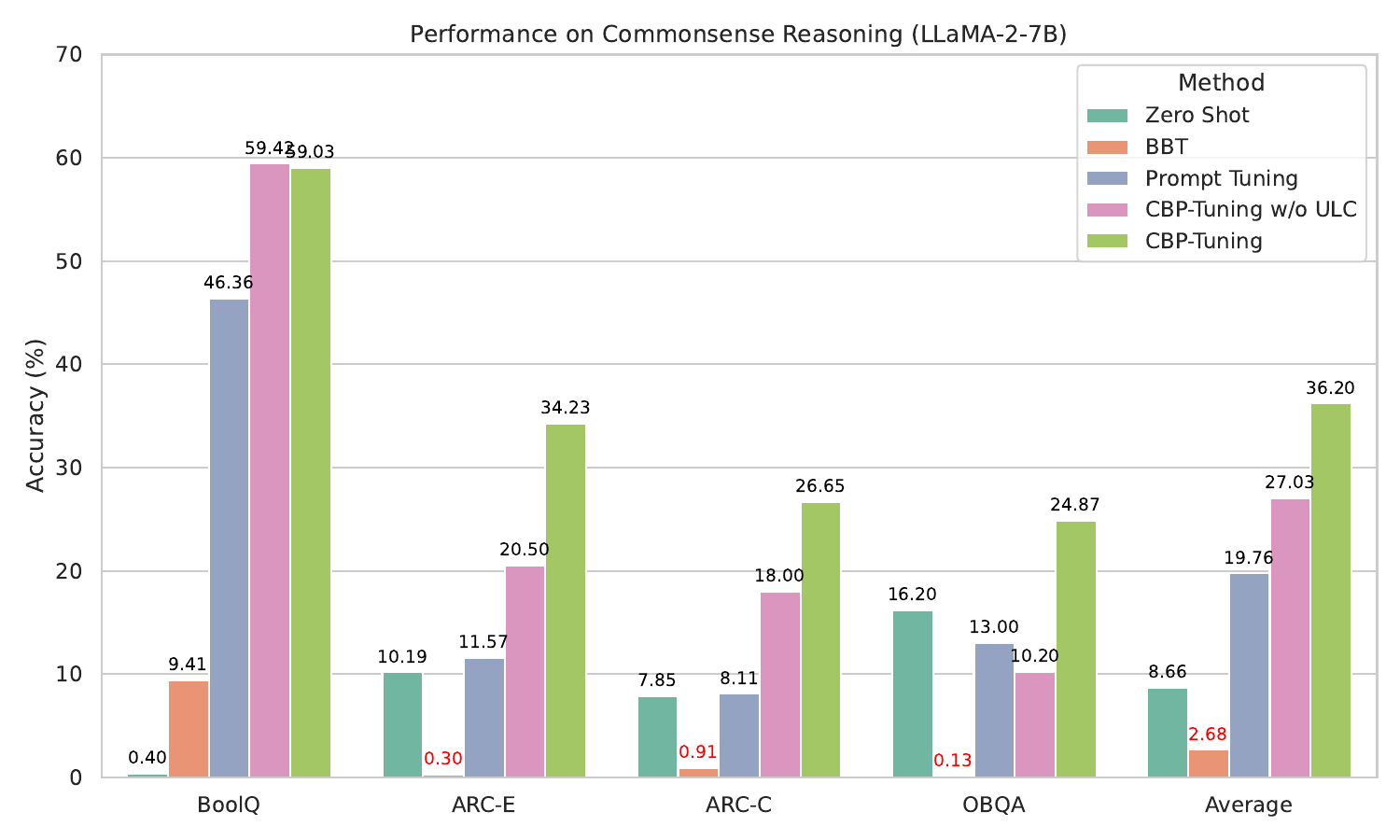}
  \caption{Results for LLaMA-2-7B model in the commonsense reasoning domain. The decrease in accuracy of BBT compared to the Zero Shot is highlighted in red.}
  \label{fig:commonsense_results}
\end{figure}

While BBT initially appears suitable for our proposed scenario, enabling local private data training without requiring model weight access, our empirical analysis reveals critical limitations. As shown in Figure \ref{fig:commonsense_results}, direct application of BBT to inject task-specific soft prompts paradoxically degrades performance in commonsense reasoning tasks, achieving only gains on BoolQ (+9.01\%) while causing significant average performance drops of 5.98 points across four datasets. This empirical evidence demonstrates that naively applying BBT for transitioning from foundation LLMs to user-customized models proves fundamentally inadequate in practical deployments. The absence of domain-specific and task-agnostic adaptation capabilities in pure BBT settings leads to unstable optimization trajectories and suboptimal task-specific performance. To bridge this gap, CBP-Tuning introduces a two-stage framework, specifically designed to overcome these architectural constraints, we introduce it in detail in the following.

\section{Method}

\subsection{Overview of CBP-Tuning}
\begin{figure*}[htbp]
  \centering
  \includegraphics[width=0.8\textwidth]{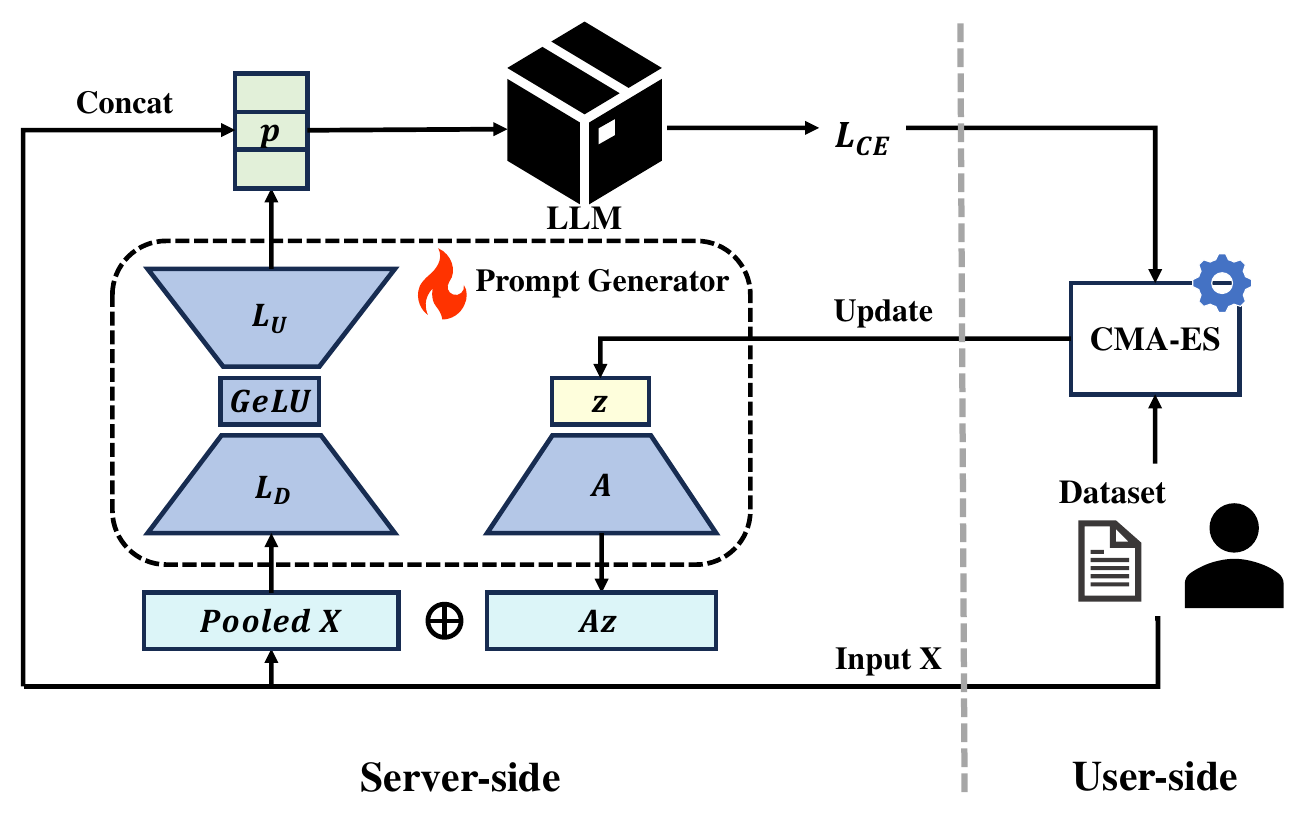}
  \caption{Illustration of Customized Black-box Prompt Tuning (CBP-Tuning). Users can perform efficient customization for each downstream task locally by using a gradient-free optimization algorithm (CMA-ES) to optimize each a low-dimensional vector $\mathbf{z}$, while the server trains a domain-specific prompt generator.}
  \label{fig:overview}
\end{figure*}

As shown in Figure \ref{fig:overview}, the workflow of CBP-Tuning includes Server-side Domain Training and User-side Local Customization. As shown in Figure \ref{fig:overview}, on the server side, we train a prompt generator on the domain dataset. Users can use the CMA-ES algorithm to optimize the vector without accessing the model weights, thus achieving efficient local customization. Next, we introduce the core component and processes of CBP-Tuning, namely prompt generator, server-side domain training, and user-side local customization.

\begin{algorithm}[tb]
\caption{The CBP-Tuning Procedure}
\label{alg:algorithm}
\begin{algorithmic}[1] 
\Require Black-box LLM $f(\cdot)$, Prompt Generator $\mathcal{G}$, Domain dataset $\{\mathcal{X}_{D}; \mathcal{Y}_{D}\}$, Training epoches $E$, Customization dataset $\{\mathcal{X}_{C}; \mathcal{Y}_{C}\}$, Budget of API calls $B$, Population size $\lambda$.
\For{$i = 1$ to $E$}
    \For{$(\mathbf{X}, \mathbf{Y}) \in \mathcal{X}_{D} \times \mathcal{Y}_{D}$}
        \State Get $\mathbf{p}$ using Eq.~\eqref{eq:prompt}
        \State Concatenate input $\mathbf{X^{'}}$ using Eq.~\eqref{eq:concat}
        \State Optimize $\mathcal{G^*}, \mathbf{z_0}$ using Eq.~\eqref{eq:training}
    \EndFor
\EndFor
\State Set $max\_iteration = B / \lambda$ for CMA-ES
\For{$j = 1$ to $max\_iteration$}
    \For{$(X, Y) \in \mathcal{X}_{C} \times \mathcal{Y}_{C}$}
        \State Get $\mathbf{p^*}$ using Eq.~\eqref{eq:cus_get_prompt}
        \State Concatenate input $X^{*}$ using Eq.~\eqref{eq:cus_concat}
        \State Optimize $\mathbf{z^*}$ using Eq.~\eqref{eq:customization}
    \EndFor
\EndFor
\end{algorithmic}
\end{algorithm}

\subsection{Prompt Generator}
LLMs are typically trained on large-scale corpora and lack the ability to handle specific user tasks. To better achieve task customization on the user side, we propose a prompt generator $\mathcal{G}$ to learn prompts that activate the corresponding domain knowledge within the LLM. 

The input $I(\mathbf{X}, \mathbf{z})$ of the prompt generator is the sum of two vectors, one is the input instruction and another is a vector for customization:
\begin{equation}
    \label{eq:input}I(\mathbf{X}, \mathbf{z}) = Pooler(Emb(\mathbf{X})) + \mathbf{Az}.
\end{equation}
On the one hand, the inputs of the model pass through the embedding layer and obtain a hidden state $\mathbf{h}\in\mathbb{R}^{l \times d}$, where $l$ is the length of the input tokens and $d$ is the model dimension. Subsequently, the hidden state $\mathbf{h}$ is reduced to a $d$-dimensional vector by mean pooling, which is a part of the input to the prompt generator. On the other hand, another part of the prompt generator's input is a vector of the same size, following BBT, we optimize the vector $\mathbf{z}\in\mathbb{R}^r$ in a smaller subspace($r \ll d$), and project $\mathbf{z}$ back to the model dimension space through a projection matrix $A\in\mathbb{R}^{d \times r}$.



The prompt generator is a lightweight module that includes a down-projection layer $\mathbf{L_{D}}\in\mathbb{R}^{m \times d}$, an activation function $GeLU(\cdot)$, and an up-projection layer $\mathbf{L_{U}}\in\mathbb{R}^{(t*d) \times m}$, which ultimately outputs a soft prompt that is concatenated with the embedding of the model input, where $t$ is the length of soft prompts. After receiving the input, the prompt generator produces the prompt:
\begin{equation}
    \label{eq:prompt}\mathbf{p} = \mathbf{L_{U}}(GeLU(\mathbf{L_{D}}(I))).
\end{equation}

Then, the prompt $\mathbf{p}$ will be prepended before the input layer, and the final input $\mathbf{X^{'}}$ is:
\begin{align}
    \label{eq:concat}\mathbf{X^{'}} = Concat(\mathbf{p}; \mathbf{X}).
\end{align}

It is important to note that our approach manipulates prompts at the input stage and concatenates them into the input layer. As a result, the prompt generator is called only once during the generation stage, unlike methods such as LoRA, which require loading modules at each layer. For users sharing the same domain, the prompt generator can be shared, supporting efficient batch processing.

\subsection{Server-side Domain Training}
To better activate the model's domain knowledge without accessing user data, we use domain data on the server side to train the prompt generator by gradient descent, allowing multiple users sharing the same domain to leverage the same prompt generator. We keep the model $f(\cdot)$ frozen and train the prompt generator $\mathcal{G}$ on the dataset from the collected domain. During the server-side domain training, our objective is to optimize:
\begin{align}
    \label{eq:training}\mathcal{G^*}, \mathbf{z_0} = {arg \min}_{\mathcal{G}\in\mathcal{G_\theta}, \mathbf{z}\in\mathcal{Z}} \mathcal{L}_{CE}(\mathbf{Y}, f(\mathbf{X^{'}})), 
\end{align}
where the input and output $\mathbf{X}\in\mathcal{X}_{D}, \mathbf{Y}\in\mathcal{Y}_{D}$ in Customization dataset $\{\mathcal{X}_{D}; \mathcal{Y}_{D}\}$, $\mathcal{G_\theta}$ is the parameter of the prompt generator, $\mathcal{Z}$ is the search space for $\mathbf{z}$, $\mathcal{L}_{CE}$ is the cross-entropy loss function. By conducting server-side domain training, we obtain a parameter generator $\mathcal{G^*}$ for a specific domain, along with an initialized low-dimensional vector $\mathbf{z}$ in this domain for subsequent user-side customization. 

\subsection{User-side Local Customization}
At the user side, we aim to customize for the user's tasks, but considering that most users lack the computational resources required for gradient descent typically needed for training, we intend to achieve this in a low-cost manner by black-box tuning. We employ CMA-ES \cite{hansen2003cmaes} to optimize $\mathbf{z_0}$ and obtain $\mathbf{z^*}$ in order to generate task-specific enhanced prompts for improved performance. Formally,
\begin{align}
    \label{eq:cus_get_prompt}\mathbf{p^*} &= \mathcal{G^*}(\mathbf{X}, \mathbf{z_0}), \\
    \label{eq:cus_concat}\mathbf{X^{*}} &= Concat(\mathbf{p^*}; \mathbf{X}), \\
    \label{eq:customization}\mathbf{z^*} &= {arg \min}_{\mathbf{z}\in\mathcal{Z}} \mathcal{L}_{CE}(\mathbf{Y}, f(\mathbf{X^{*}})), 
\end{align}
where the input and output are  $\mathbf{X}\in\mathcal{X}_{C}$ and $\mathbf{Y}\in\mathcal{Y}_{C}$ respectively, and $\{\mathcal{X}_{C}; \mathcal{Y}_{C}\}$ is the customization dataset.

Similarly, the derivative-free optimization algorithm is also guided by the cross-entropy loss $\mathcal{L}_{CE}$ to optimize the direction. During the local customization stage, users cannot access the model weight, and the model and parameter generator are kept fixed. 

So far, users simply require the server to transmit a domain-specific prompt generator to implement local customization for each downstream task. On the server side, we train $\mathbf{L_{U}}, \mathbf{L_{D}}, \mathbf{A}, \mathbf{z}$, where the total number of parameters is $m \times d + m \times l \times d + r \times d + r$ required to be stored and sent, while the user side only needs to save $s \times r$ parameters, where $s$ is the number of downstream tasks and $r$ is a low dimension of the vector $\mathbf{z}$.

\begin{table*}[t]
    \centering
    \resizebox{\textwidth}{!}{
    \begin{tabular}{lcccccccclcccc}
        \toprule
        \multirow{2}{*}{Model} & \multirow{2}{*}{Setting} 
        & \multicolumn{7}{c}{Medical Domain}
        & & \multicolumn{4}{c}{Financial Domain} \\
        \cmidrule{3-9} \cmidrule{11-14}
        & & CK & MG & Anatomy & PM & CB & CM & Med Avg & & FIQA\_SA & TFNS & FPB & Fin Avg \\
        \hline
        \multirow{4}{*}{\textbf{LLaMA-2-7B}} 
        & Zero Shot & 2.26 & 5.00 & 0.74 & 0.00 & 0.00 & 1.16 & 1.53 & & 23.93 & 13.23 & 3.75 & 13.64 \\
        & Prompt Tuning & 19.62 & 20.00 & 11.85 & 19.12 & 20.14 & 16.76 & 17.92 & & 22.65 & 11.81 & 10.51 & 14.99 \\
        & CBP-Tuning w/o ULC & 28.30 & 26.00 & 25.19 & 11.76 & 29.17 & \textbf{26.59} & 24.50 & & 59.83 & 25.42 & 43.02 & 42.76 \\
        & CBP-Tuning & \textbf{34.21} & \textbf{37.00} & \textbf{29.63} & \textbf{21.20} & \textbf{29.40} & 21.97 & \textbf{28.90} & & \textbf{63.25} & \textbf{34.17} & \textbf{45.32} & \textbf{47.58} \\
        \hline
        \multirow{4}{*}{\textbf{Qwen-2.5-3B}} 
        & Zero Shot & 39.25 & 49.00 & 49.63 & 20.59 & 36.81 & 39.31 & 39.10 & & 66.67 & 38.65 & 77.96 & 61.09 \\
        & Prompt Tuning & 65.66 & 71.00 & 54.81 & 37.87 & 67.36 & 60.12 & 59.47 & & 58.12 & 69.26 & \textbf{83.97} & 70.45 \\
        & CBP-Tuning w/o ULC & 67.92 & \textbf{74.00} & \textbf{59.26} & 66.54 & \textbf{74.31} & 61.85 & 67.31 & & 62.39 & 69.85 & 72.44 & 68.53 \\
        & CBP-Tuning & \textbf{68.55} & \textbf{74.00} & \textbf{59.26} & \textbf{69.49} & \textbf{74.31} & \textbf{63.39} & \textbf{68.17} & & \textbf{72.22} & \textbf{70.45} & 77.80 & \textbf{73.49} \\
        \hline
        \multirow{4}{*}{\textbf{LLaMA-2-13B}} 
        & Zero Shot & 17.36 & 13.00 & 13.33 & 0.74 & 15.28 & 12.14 & 11.97 & & 68.38 & 22.82 & 27.69 & 39.63 \\
        & Prompt Tuning & 18.87 & 12.00 & 9.63 & 23.16 & 14.58 & 23.12 & 16.89 & & 44.44 & \textbf{65.49} & 65.81 & 58.58 \\
        & CBP-Tuning w/o ULC & 61.13 & 46.00 & 34.07 & 47.06 & 55.56 & \textbf{48.55} & 48.73 & & 65.38 & 53.69 & 63.83 & 60.96 \\
        & CBP-Tuning & \textbf{61.51} & \textbf{46.00} & \textbf{42.96} & \textbf{48.28} & \textbf{56.25} & 47.40 & \textbf{50.40} & & \textbf{75.64} & 58.04 & \textbf{71.73} & \textbf{68.47} \\
        \bottomrule
    \end{tabular}
    }
    \caption{Performance (\%) of LLaMA-2-7B, Qwen-2.5-3B, and LLaMA-2-13B models across medical and financial domains. `Med Avg` and `Fin Avg` indicate average scores over each domain's subtasks.}
    \label{tab:merged_medical_financial}
\end{table*}

\section{Experiments}

\subsection{Setup}
\label{sec:setup}
\paragraph{Models and Datasets.} 
We implement experiments on LLaMA-2-7B, LLaMA-2-13B \cite{touvron2023llama} and Qwen-2.5-3B \cite{yang2024qwen2} models. 

To simulate the user's private data, we perform a complete separation between the domain dataset used for the Server-side Domain Training (SDT) stage and the customization dataset used for the User-side Local Customization (ULC) stage within the same domain. Comprehensive evaluations are conducted on both a relatively general commonsense reasoning domain and a more specialized medical domain. 

For the commonsense reasoning domain, the domain dataset in the SDT stage is composed of training sets from PIQA \cite{bisk2020piqa}, SIQA \cite{sap2019socialiqa}, HellaSwag \cite{zellers2019hellaswag}, and WinoGrande \cite{sakaguchi2021winogrande}. The customization dataset in the ULC stage includes BoolQ \cite{clark2019boolq}, the challenge set ARC-Challenge and easy set ARC-Easy of ARC \cite{clark2018arc}, and OpenBookQA \cite{mihaylov2018openbookqa}. For each task in the customization dataset, we randomly select 16 shots from its training set. 

For the medical domain, we construct a domain dataset by combining a subset of approximately 6,000 samples from a biomedical instruction-following dataset Medical Meadow \cite{han2023medalpaca}, and the first 100,000 samples from Medical Multiple-Choice Question Answering (MedMCQA) \cite{pal2022medmcqa}. This domain dataset is used for the SDT stage to enable LLMs to learn how to handle complex medical instructions while also answering medical-related questions. For the customization dataset, we use 6 subsets of MMLU \cite{hendrycks2020mmlu} that are most relevant to medical knowledge, specifically anatomy, clinical knowledge, college medicine, medical genetics, college biology, and professional medicine. In the ULC phase, we utilize the development sets of these subsets and report their average performance. 

For the financial domain, we selected a mixture of the first 10k samples from the sentiment training set in FinGPT's \cite{yang2023fingpt} instruction-tuning dataset and the first 1k samples from the FiQA\_QA \cite{cheng2024fiqa} dataset as the training set for the SDT stage. For testing, we employed three financial sentiment analysis datasets: Twitter Financial News Sentiment, FiQA\_SA \cite{cheng2024fiqa}, and a subset of the Financial Phrase Bank \cite{Malo2014fpb} containing only samples with 100\% annotator confidence. In the ULC stage, we randomly selected 16 samples for optimization.

We conduct experiments under three different random seeds in the three domains. Our evaluation metrics follow the setup of LLM-Adapters \cite{hu2023llm-adapters}. The detailed setup is provided in Appendix \ref{sec:appendix-a}. 

\paragraph{Baselines.} 
We compare our proposed method with the following four baselines. (1) \textbf{Zero Shot:} We directly use the alpaca-style \cite{alpaca} prompt template to test the model on the given tasks. (2) \textbf{Prompt Tuning:} Following \cite{lester2021prompt-tuning}, we freeze the model and only train the prompts prepended to the input embedding. (3) \textbf{CBP-Tuning w/o ULC:} We train on a mixed domain dataset, i.e., server-side domain training for CBP-Tuning without further user local customization, then we can implement the complete CBP-Tuning process based on this. (4) \textbf{CBP-Tuning:} We use the CMA-ES algorithm to optimize without accessing the model. In the CMA-ES optimization process, we input a batch of training data together and optimize using the average cross-entropy loss for the answer output part. 
To ensure a fair comparison, Prompt Tuning, CBP-Tuning, and its w/o ULC baseline are fine-tuned using the same domain dataset within the same domain. The prompt template is presented in Appendix \ref{sec:appendix-a-prompt-template}.
(5) \textbf{LoRA:} A widely-used PEFT method \cite{hu2021lora} that freezes pre-trained model weights and injects trainable, low-rank matrices into the Transformer layers for efficient adaptation. (6) \textbf{P-Tuning:} A prompt-based method \cite{liu2021p-tuning} that keeps the language model frozen and optimizes continuous prompt embeddings via a small prompt encoder. The detailed experimental comparison with these additional baselines is provided in Appendix \ref{sec:appendix-b-baselines}.

\paragraph{Implementation.}
 For Prompt Tuning, we evaluate each task on the customization dataset after training on the domain dataset. For the SDT stage of CBP-Tuning and its variants, the intermediate dimension $m$ of the prompt generator is set to 256, and the subspace dimension $r$ of vector $\mathbf{z}$ is set to 500. All the above methods keep the model weights frozen during training, and the length of the soft prompt concatenated before the user input embedding is 10. For the ULC stage of CBP-Tuning, the customization dataset size, budget of black-box LLM calls $B$, population size $\lambda$, and $\sigma$ are set to \{16, 300, 30, 0.01\} in the commonsense reasoning domain, respectively. The hyperparameters during training and the optimization parameters for other domains in the ULC phase are detailed in the Appendix \ref{sec:appendix-a-hyperparameters}.

\subsection{Main Results}
\label{sec:main_results}

We demonstrate the results of baselines on four commonsense reasoning datasets based on the LLaMA-2-7B model and six medical datasets based on all models in Figure \ref{fig:commonsense_results} and Table \ref{tab:merged_medical_financial} respectively. 

\paragraph{Comparison on Zero Shot. }
Across all domains, the untrained zero-shot approach demonstrates the weakest performance among all baselines. Notably, its performance is significantly poorer in the medical and financial domains which require domain-specific knowledge compared to the more general commonsense reasoning domain. While trained methods including Prompt Tuning, CBP-Tuning, and its w/o ULC variant all achieve substantial performance improvements, we observe slight performance degradation for both Prompt Tuning and CBP-Tuning w/o ULC on the OpenBookQA task. This discrepancy may be attributed to potential conflicts between the training data used in the SDT phase and the OpenBookQA task. Importantly, the CBP-Tuning framework achieves comprehensive performance gains across all evaluated datasets.

\paragraph{Comparison on Prompt Tuning. }
Our experiments reveal that CBP-Tuning and its w/o ULC variant substantially outperform Prompt Tuning, with CBP-Tuning achieving average performance advantages of 16.44 points over Prompt Tuning in commonsense reasoning, and 32.59 points in the financial domain using LLaMA-2-7B. For LLaMA-2-13B, CBP-Tuning boosts the Med Avg from 16.89 (Prompt Tuning) to 50.40 and the Fin Avg from 58.58 to 68.47, showcasing significant gains. Importantly, the full CBP-Tuning framework demonstrates comprehensive performance gains over the Prompt Tuning baseline across all benchmarks. This evidences that through its two-stage optimization, CBP-Tuning not only addresses privacy preservation requirements in client-server deployments but also delivers superior performance compared to directly serving server-side fine-tuned Prompt Tuning models to end-users.

\paragraph{Comparison on w/o User-side Local Customization. }
Our comparative analysis of CBP-Tuning and its w/o ULC variant illustrates that the ULC phase generally delivers substantial performance improvements, yielding average gains of 9.17 points in the commonsense reasoning domain, 4.40, 0.86 and 1.67 points in the medical domain using LLaMA-2-7B, Qwen-2.5-3B and LLaMA-2-13B respectively. In the financial domain, the ULC phase provides average gains of 4.82 points for LLaMA-2-7B, 4.96 points for Qwen-2.5-3B, and 7.51 points for LLaMA-2-13B. However, we observe localized performance degradation on BoolQ and MMLU Clinical Medicine (CM) tasks with LLaMA-2-7B model. The BoolQ decline may stem from its unique yes/no answer format diverging from other tasks' option format, while MMLU CM's challenges originate from its benchmark characteristics: abbreviated development sets (used for the ULC) coupled with test set questions exceeding 480 words in 8.1\% cases. Furthermore, as shown in the Appendix \ref{sec:appendix-b-analysis}, we examine the final loss of the LLaMA-2-7B model during the ULC phase in the medical domain and found that the loss for the poorly performing CM task was 1–3 orders of magnitude larger than that of other tasks. Crucially, the ULC stage demonstrates task-agnostic effectiveness in boosting model performance, achieving these gains through gradient-free local optimization that preserves privacy constraints.

\begin{figure}[t]
  \centering
  \includegraphics[width=0.4\textwidth]{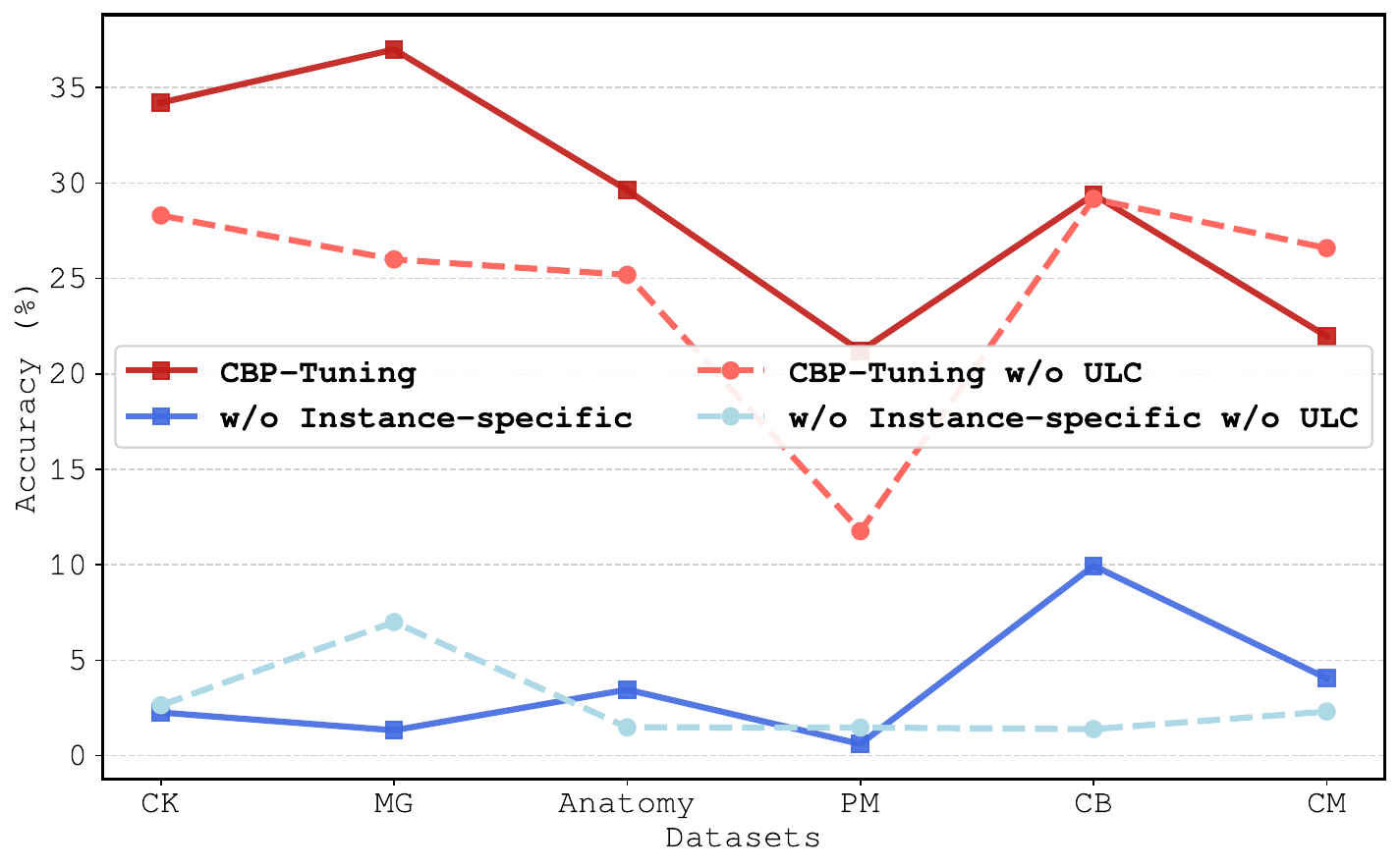}
  \caption{Results for ablation study of instance-specific information in the medical domain using LLaMA-2-7B.
  }
  \label{fig:pic_ablation_no_ins}
\end{figure}

\subsection{Ablation Study of Instance-specific Information}
To verify the effectiveness of introducing instance-specific information into the prompt generator, we set up an ablation version without instance-specific information. We compare the setting without mean pooled instance embedding in the medical domain, and all other settings are the same as CBP-Tuning. We can see that the performance of the w/o Instance-specific setting and its w/o ULC variant (represented by the blue solid and dashed lines in Figure \ref{fig:pic_ablation_no_ins}, respectively) is significantly weaker than the standard CBP-Tuning and its w/o ULC variant (represented by the red solid and dashed lines in Figure \ref{fig:pic_ablation_no_ins}, respectively). Specifically, the average performance across the six datasets in the medical domain shows that the w/o Instance-specific version drops by 25.29 points, and its w/o ULC variant drops by 21.28 points. 

\begin{figure}[t]
  \centering
  \includegraphics[width=0.4\textwidth]{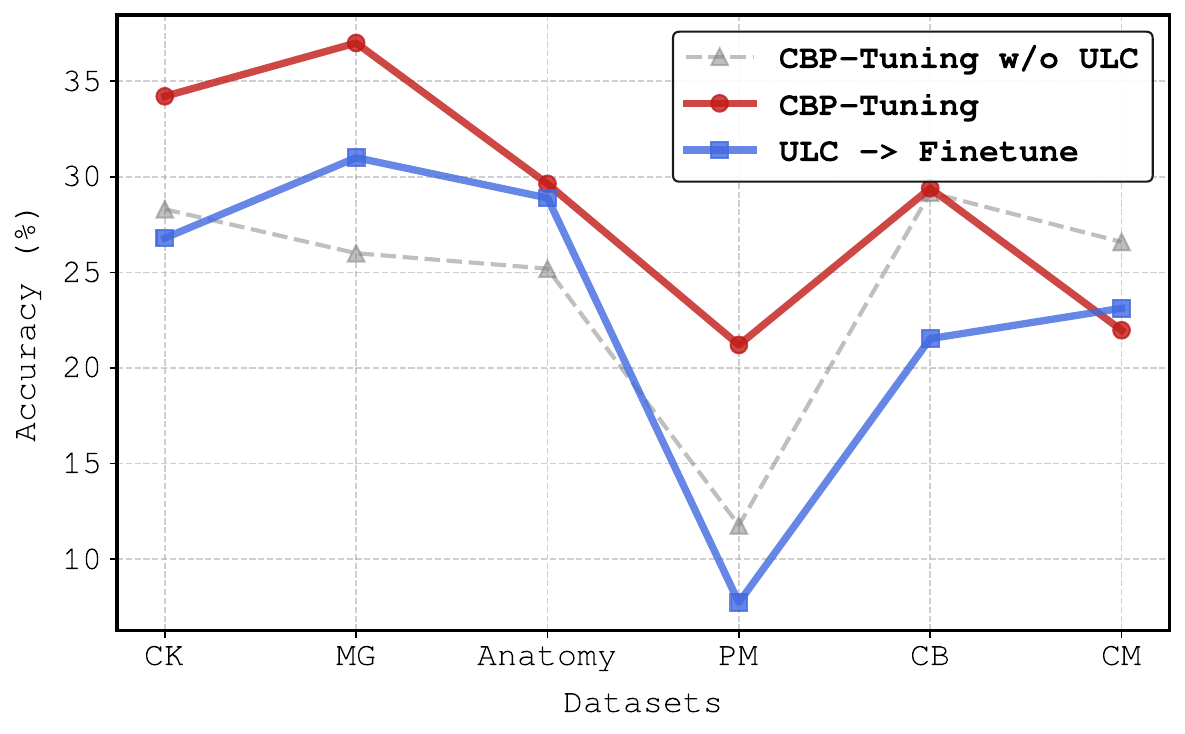}
  \caption{Results for optimization methods analysis in the medical domain using LLaMA-2-7B. 
  }
  \label{fig:pic_ablation_optimization}
\end{figure}

\subsection{Different Optimization Methods Analysis}
As previously mentioned, implementing privacy protection for both the server and user sides in our framework requires the execution of ULC in the second stage. To investigate the impact of replacing the user local customization phase with fine-tuning, we performed a few-shot fine-tuning version and explored the effects of further fine-tuning using the customization dataset (which, in the medical domain, corresponds to the development set of each task). As shown in the Figure \ref{fig:pic_ablation_optimization}, the blue line, red line, and gray dashed line represent the few-shot fine-tuning setting, the standard CBP-Tuning, and its w/o ULC variant, respectively. It is evident that using ULC in the second stage, i.e., gradient-free optimization, consistently outperforms gradient-based fine-tuning. CBP-Tuning performs slightly worse than its few-shot fine-tuning variant only on the CM dataset, with an average improvement of 5.72 points across six medical datasets. In contrast, the few-shot fine-tuning setting shows performance improvements over CBP-Tuning w/o ULC only on the MG and Anatomy datasets, while it leads to a performance decline on the other datasets. We suspect two potential reasons: (1) While gradient-based optimization tends to overfit the training data, CMA-ES is inclined to discover superior solutions owing to its exploration mechanism. (2) The server-side training instills a structured instruction-following format in the model. However, certain downstream tasks may have unique characteristics that create a mismatch with this pre-trained format. This discrepancy makes the optimization landscape for the vector $\mathbf{z}$ particularly challenging. For tasks where this gap is significant (e.g., MMLU Clinical Medicine), the CMA-ES algorithm may fail to converge to an effective solution, a difficulty empirically evidenced by the high final loss values for such tasks, as detailed in Appendix~\ref{tab:cm_optimization_losses}.

\subsection{Efficiency Analysis}
\label{sec:efficiency_analysis}

In Table \ref{tab:training_time}, we show the training time of user-side local customization on four commonsense reasoning datasets using LLaMA-2-7B model. On a single NVIDIA A100 40GB GPU, the optimization can be completed in just 93.5 seconds across these datasets, with the shortest time recorded at 54 seconds on the BoolQ dataset. In contrast, on average, fine-tuning through Prompt Tuning typically demands more than ten minutes. In commonsense reasoning domain's experimental setup, the user-side optimization only requires calling the API 300 times and performing 10 iterations of the CMA-ES algorithm. As shown in Table \ref{tab:gpu_memory}, when using a batch size of 16, our method demonstrates significantly lower GPU memory consumption compared to Prompt tuning. Notably, since CMA-ES optimization does not rely on gradients, the entire optimization process can be performed on the CPU. Overall, our proposed approach enables users to efficiently and locally customize the model in an exceedingly short duration.

\begin{table}[t]
    \centering
    \resizebox{\columnwidth}{!}{
    \begin{tabular}{lcccc}
        \toprule
         \textbf{Method} & \textbf{OBQA} & \textbf{BoolQ} & \textbf{ARC-E} & \textbf{ARC-C} \\
         \midrule
         CBP-Tuning & 85s & 54s & 101s & 109s \\ 
         \cmidrule(lr){2-5}
         Prompt Tuning & \multicolumn{4}{c}{\textit{$>$ 10 mins on each dataset}} \\
        \bottomrule
    \end{tabular}
    }
    \caption{Comparison of user-side customization time for CBP-Tuning versus full fine-tuning time for Prompt Tuning in the commonsense reasoning domain. All timings were measured on a single NVIDIA A100 40GB GPU using LLaMA-2-7B.}
    \label{tab:training_time}
\end{table}

\begin{table}[t]
    \centering
    \resizebox{\columnwidth}{!}{
    \begin{tabular}{lcccc}
        \toprule
        \textbf{Method} & \textbf{OBQA} & \textbf{BoolQ} & \textbf{ARC-E} & \textbf{ARC-C} \\
        \midrule
        CBP-Tuning & 16854MB & 15668MB & 17460MB & 17864MB \\
        Prompt Tuning & 28572MB & 21482MB & 31712MB & 33544MB \\
        \bottomrule
    \end{tabular}
    }
    \caption{GPU memory usage in the commonsense reasoning domain on LLaMA-2-7B.}
    \label{tab:gpu_memory}
\end{table}

\section{Related Work}

\subsection{Prompt-based Learning}

Prompt-based learning is a type of method that inserts soft prompts into the input or hidden states of the model. These soft prompts are highly flexible and adaptable during fine-tuning, as they can be optimized in continuous spaces to fit specific tasks. Methods like Prompt Tuning \cite{lester2021prompt-tuning} and P-tuning \cite{liu2021p-tuning} incorporate soft prompts into the input layer of the model. Other approaches, such as DePT \cite{shi2023dept}, Prefix Tuning \cite{li2021prefix}, and P-tuning v2 \cite{liu2021p-tuningv2}, add trainable prompts to the keys and values matrices across all layers. These methods focus primarily on task-specific prompt tuning. Additionally, there are methods that integrate instance-specific information. For example, IDPG \cite{wu2022idpg} uses a parameterized hypercomplex multiplication prompt generator to produce soft prompts tailored for each instance. LPT \cite{liu2022lateprompttuning} inserts instance-aware prompts into an intermediate layer of the pre-trained model, rather than the input layer or all layers. We observe that these two types of methods for improving prompt tuning performance decouple task-specific information from instance-specific information and reserve a task vector suitable for local customization at the input of the prompt generator for two-stage optimization.

\subsection{Black-box Optimization for LLMs} 

Black-Box Tuning \cite{sun2022black-box} innovatively uses a derivative-free optimization method to learn prompts in the new scenario called Language-Model-as-a-Service (LMaaS). BBTv2 \cite{sun2022bbtv2}, as an improved version, prepends continuous prompts to every layer of the model and optimizes the prompts at different layers alternately. InstructZero \cite{chen2023instructzero} employs Bayesian optimization to learn the continuous prompts inserted into open-source LLMs to generate discrete instructions for the black-box LLMs. In addition to using gradient-free optimization algorithms to optimize soft prompts, there are also some recent studies showing that it is possible to optimize black-box large models using white-box models with varying degrees of transparency. Offsite Tuning \cite{xiao2023offsite} allows users to adapt to downstream tasks with the help of a lightweight adapter and a compressed emulator without accessing the full model.  BBox-Adapter \cite{sun2024bbox} adapts black-box LLMs to specific tasks without accessing internal parameters or output probabilities by training a small adapter model with ranking-based NCE loss and online adaptation. Proxy-tuning \cite{liu2024proxytuning} tunes a smaller LM, applying the difference between the predictions of the small tuned and untuned LMs to shift the direction of tuning for the larger untuned model at the cost of exposing logits. Based on the observation that directly applying black-box optimization to LLMs for certain tasks often results in insufficient performance \cite{zheng2024bpt-subspace}, we only need access and modification after the embedding layer to locally customize black-box LLMs driven by user needs.

\section{Conclusion}

We present CBP-Tuning, a lightweight and customized fine-tuning framework for black-box large language models. Our approach enables efficient local customization through a black-box optimization method that balances the burden between the server and the user. CBP-Tuning allows users to achieve cost-effective customization for various downstream tasks without compromising privacy. We anticipate that future research can extend CBP-Tuning to a broader range of domains and models.

\section*{Limitations}

While CBP-Tuning demonstrates promising privacy-preserving capabilities and competitive performance across both target domains, future efforts could focus on two aspects. First, our evaluation was limited to models up to the 13B parameter scale; its efficacy on significantly larger models (e.g., 70B+) remains unexplored due to computational constraints. Second, the user-side customization stage led to performance degradation on specific tasks like BoolQ and MMLU Clinical Medicine. Future efforts could explore adaptive optimization strategies or methods to enhance the robustness of ULC across a more diverse range of task structures.

\section*{Ethics Statement}

Our work introduces CBP-Tuning, a framework designed for efficient and privacy-preserving local customization of large language models. All datasets used in our study for training and evaluation are publicly available and sourced from established academic benchmarks. No proprietary or sensitive user data was involved in our experiments. Our method is purely algorithmic and is not designed to inherently create or amplify harmful social biases.

\section*{Acknowledgement}

We thank the anonymous reviewers for their insightful feedback, which greatly improves our paper. This work is supported by the National Natural Science Foundation of China (No. 62472419, 62472420).

\bibliography{custom}

\begin{thebibliography}{45}
\providecommand{\natexlab}[1]{#1}

\bibitem[{Bisk et~al.(2020)Bisk, Zellers, Gao, Choi et~al.}]{bisk2020piqa}
Yonatan Bisk, Rowan Zellers, Jianfeng Gao, Yejin Choi, et~al. 2020.
\newblock Piqa: Reasoning about physical commonsense in natural language.
\newblock In \emph{Proceedings of the AAAI conference on artificial intelligence}, volume~34, pages 7432--7439.

\bibitem[{Brown et~al.(2020)Brown, Mann, Ryder, Subbiah, Kaplan, Dhariwal, Neelakantan, Shyam, Sastry, Askell et~al.}]{gpt}
Tom Brown, Benjamin Mann, Nick Ryder, Melanie Subbiah, Jared~D Kaplan, Prafulla Dhariwal, Arvind Neelakantan, Pranav Shyam, Girish Sastry, Amanda Askell, et~al. 2020.
\newblock Language models are few-shot learners.
\newblock \emph{Advances in neural information processing systems}, 33:1877--1901.

\bibitem[{Chen et~al.(2023)Chen, Chen, Goldstein, Huang, and Zhou}]{chen2023instructzero}
Lichang Chen, Jiuhai Chen, Tom Goldstein, Heng Huang, and Tianyi Zhou. 2023.
\newblock Instructzero: Efficient instruction optimization for black-box large language models.
\newblock \emph{arXiv preprint arXiv:2306.03082}.

\bibitem[{Cheng et~al.(2024)Cheng, Huang, and Wei}]{cheng2024fiqa}
Daixuan Cheng, Shaohan Huang, and Furu Wei. 2024.
\newblock Adapting large language models via reading comprehension.
\newblock In \emph{ICLR}.

\bibitem[{Clark et~al.(2019)Clark, Lee, Chang, Kwiatkowski, Collins, and Toutanova}]{clark2019boolq}
Christopher Clark, Kenton Lee, Ming-Wei Chang, Tom Kwiatkowski, Michael Collins, and Kristina Toutanova. 2019.
\newblock Boolq: Exploring the surprising difficulty of natural yes/no questions.
\newblock \emph{arXiv preprint arXiv:1905.10044}.

\bibitem[{Clark et~al.(2018)Clark, Cowhey, Etzioni, Khot, Sabharwal, Schoenick, and Tafjord}]{clark2018arc}
Peter Clark, Isaac Cowhey, Oren Etzioni, Tushar Khot, Ashish Sabharwal, Carissa Schoenick, and Oyvind Tafjord. 2018.
\newblock Think you have solved question answering? try arc, the ai2 reasoning challenge.
\newblock \emph{arXiv preprint arXiv:1803.05457}.

\bibitem[{Feng et~al.(2025)Feng, Shen, Gu, Zhao, Fu, Lin, and Wang}]{feng2025divemoediversityenhancedreconstruction}
Yuchen Feng, Bowen Shen, Naibin Gu, Jiaxuan Zhao, Peng Fu, Zheng Lin, and Weiping Wang. 2025.
\newblock \href {https://arxiv.org/abs/2506.09351} {Dive into moe: Diversity-enhanced reconstruction of large language models from dense into mixture-of-experts}.
\newblock \emph{Preprint}, arXiv:2506.09351.

\bibitem[{Gu et~al.(2025{\natexlab{a}})Gu, Fu, Liu, Ma, Lin, and Wang}]{gu2025adaptoncethriveupdates}
Naibin Gu, Peng Fu, Xiyu Liu, Ke~Ma, Zheng Lin, and Weiping Wang. 2025{\natexlab{a}}.
\newblock \href {https://arxiv.org/abs/2506.06844} {Adapt once, thrive with updates: Transferable parameter-efficient fine-tuning on evolving base models}.
\newblock \emph{Preprint}, arXiv:2506.06844.

\bibitem[{Gu et~al.(2024)Gu, Fu, Liu, Shen, Lin, and Wang}]{gu-etal-2024-light}
Naibin Gu, Peng Fu, Xiyu Liu, Bowen Shen, Zheng Lin, and Weiping Wang. 2024.
\newblock \href {https://aclanthology.org/2024.findings-acl.447} {Light-{PEFT}: Lightening parameter-efficient fine-tuning via early pruning}.
\newblock In \emph{Findings of the Association for Computational Linguistics ACL 2024}, pages 7528--7541, Bangkok, Thailand and virtual meeting. Association for Computational Linguistics.

\bibitem[{Gu et~al.(2025{\natexlab{b}})Gu, Zhang, Liu, Fu, Lin, Wang, Sun, Wu, Wang, and Wang}]{DBLP:conf/acl/Gu0000WS00025}
Naibin Gu, Zhenyu Zhang, Xiyu Liu, Peng Fu, Zheng Lin, Shuohuan Wang, Yu~Sun, Hua Wu, Weiping Wang, and Haifeng Wang. 2025{\natexlab{b}}.
\newblock \href {https://aclanthology.org/2025.acl-long.582/} {Beamlora: Beam-constraint low-rank adaptation}.
\newblock In \emph{Proceedings of the 63rd Annual Meeting of the Association for Computational Linguistics (Volume 1: Long Papers), {ACL} 2025, Vienna, Austria, July 27 - August 1, 2025}, pages 11871--11883. Association for Computational Linguistics.

\bibitem[{Gururajan et~al.(2024)Gururajan, Lopez-Cuena, Bayarri-Planas, Tormos, Hinjos, Bernabeu-Perez, Arias-Duart, Martin-Torres, Urcelay-Ganzabal, Gonzalez-Mallo et~al.}]{gururajan2024aloe}
Ashwin~Kumar Gururajan, Enrique Lopez-Cuena, Jordi Bayarri-Planas, Adrian Tormos, Daniel Hinjos, Pablo Bernabeu-Perez, Anna Arias-Duart, Pablo~Agustin Martin-Torres, Lucia Urcelay-Ganzabal, Marta Gonzalez-Mallo, et~al. 2024.
\newblock Aloe: A family of fine-tuned open healthcare llms.
\newblock \emph{arXiv preprint arXiv:2405.01886}.

\bibitem[{Han et~al.(2023)Han, Adams, Papaioannou, Grundmann, Oberhauser, L{\"o}ser, Truhn, and Bressem}]{han2023medalpaca}
Tianyu Han, Lisa~C Adams, Jens-Michalis Papaioannou, Paul Grundmann, Tom Oberhauser, Alexander L{\"o}ser, Daniel Truhn, and Keno~K Bressem. 2023.
\newblock Medalpaca--an open-source collection of medical conversational ai models and training data.
\newblock \emph{arXiv preprint arXiv:2304.08247}.

\bibitem[{Han et~al.(2024)Han, Gao, Liu, Zhang et~al.}]{peft-survey}
Zeyu Han, Chao Gao, Jinyang Liu, Sai~Qian Zhang, et~al. 2024.
\newblock Parameter-efficient fine-tuning for large models: A comprehensive survey.
\newblock \emph{arXiv preprint arXiv:2403.14608}.

\bibitem[{Hansen et~al.(2003)Hansen, M{\"u}ller, and Koumoutsakos}]{hansen2003cmaes}
Nikolaus Hansen, Sibylle~D M{\"u}ller, and Petros Koumoutsakos. 2003.
\newblock Reducing the time complexity of the derandomized evolution strategy with covariance matrix adaptation (cma-es).
\newblock \emph{Evolutionary computation}, 11(1):1--18.

\bibitem[{He et~al.(2021)He, Zhou, Ma, Berg{-}Kirkpatrick, and Neubig}]{DBLP:journals/corr/abs-2110-04366}
Junxian He, Chunting Zhou, Xuezhe Ma, Taylor Berg{-}Kirkpatrick, and Graham Neubig. 2021.
\newblock \href {https://arxiv.org/abs/2110.04366} {Towards a unified view of parameter-efficient transfer learning}.
\newblock \emph{CoRR}, abs/2110.04366.

\bibitem[{Hendrycks et~al.(2020)Hendrycks, Burns, Basart, Zou, Mazeika, Song, and Steinhardt}]{hendrycks2020mmlu}
Dan Hendrycks, Collin Burns, Steven Basart, Andy Zou, Mantas Mazeika, Dawn Song, and Jacob Steinhardt. 2020.
\newblock Measuring massive multitask language understanding.
\newblock \emph{arXiv preprint arXiv:2009.03300}.

\bibitem[{Houlsby et~al.(2019)Houlsby, Giurgiu, Jastrzebski, Morrone, De~Laroussilhe, Gesmundo, Attariyan, and Gelly}]{houlsby2019adapter}
Neil Houlsby, Andrei Giurgiu, Stanislaw Jastrzebski, Bruna Morrone, Quentin De~Laroussilhe, Andrea Gesmundo, Mona Attariyan, and Sylvain Gelly. 2019.
\newblock Parameter-efficient transfer learning for nlp.
\newblock In \emph{International conference on machine learning}, pages 2790--2799. PMLR.

\bibitem[{Hu et~al.(2021)Hu, Shen, Wallis, Allen-Zhu, Li, Wang, Wang, and Chen}]{hu2021lora}
Edward~J Hu, Yelong Shen, Phillip Wallis, Zeyuan Allen-Zhu, Yuanzhi Li, Shean Wang, Lu~Wang, and Weizhu Chen. 2021.
\newblock Lora: Low-rank adaptation of large language models.
\newblock \emph{arXiv preprint arXiv:2106.09685}.

\bibitem[{Hu et~al.(2023)Hu, Wang, Lan, Xu, Lim, Bing, Xu, Poria, and Lee}]{hu2023llm-adapters}
Zhiqiang Hu, Lei Wang, Yihuai Lan, Wanyu Xu, Ee-Peng Lim, Lidong Bing, Xing Xu, Soujanya Poria, and Roy Lee. 2023.
\newblock Llm-adapters: An adapter family for parameter-efficient fine-tuning of large language models.
\newblock In \emph{Proceedings of the 2023 Conference on Empirical Methods in Natural Language Processing}, pages 5254--5276.

\bibitem[{Lester et~al.(2021)Lester, Al-Rfou, and Constant}]{lester2021prompt-tuning}
Brian Lester, Rami Al-Rfou, and Noah Constant. 2021.
\newblock The power of scale for parameter-efficient prompt tuning.
\newblock In \emph{Proceedings of the 2021 Conference on Empirical Methods in Natural Language Processing}. Association for Computational Linguistics.

\bibitem[{Li and Liang(2021)}]{li2021prefix}
Xiang~Lisa Li and Percy Liang. 2021.
\newblock Prefix-tuning: Optimizing continuous prompts for generation.
\newblock In \emph{Proceedings of the 59th Annual Meeting of the Association for Computational Linguistics and the 11th International Joint Conference on Natural Language Processing (Volume 1: Long Papers)}, pages 4582--4597.

\bibitem[{Liu et~al.(2024)Liu, Han, Wang, Tsvetkov, Choi, and Smith}]{liu2024proxytuning}
Alisa Liu, Xiaochuang Han, Yizhong Wang, Yulia Tsvetkov, Yejin Choi, and Noah~A Smith. 2024.
\newblock Tuning language models by proxy.
\newblock \emph{arXiv preprint arXiv:2401.08565}.

\bibitem[{Liu et~al.(2022)Liu, Sun, Huang, and Qiu}]{liu2022lateprompttuning}
Xiangyang Liu, Tianxiang Sun, Xuan-Jing Huang, and Xipeng Qiu. 2022.
\newblock Late prompt tuning: A late prompt could be better than many prompts.
\newblock In \emph{Findings of the Association for Computational Linguistics: EMNLP 2022}, pages 1325--1338.

\bibitem[{Liu et~al.(2021{\natexlab{a}})Liu, Ji, Fu, Tam, Du, Yang, and Tang}]{liu2021p-tuningv2}
Xiao Liu, Kaixuan Ji, Yicheng Fu, Weng~Lam Tam, Zhengxiao Du, Zhilin Yang, and Jie Tang. 2021{\natexlab{a}}.
\newblock P-tuning v2: Prompt tuning can be comparable to fine-tuning universally across scales and tasks.
\newblock \emph{arXiv preprint arXiv:2110.07602}.

\bibitem[{Liu et~al.(2021{\natexlab{b}})Liu, Zheng, Du, Ding, Qian, Yang, and Tang}]{liu2021p-tuning}
Xiao Liu, Yanan Zheng, Zhengxiao Du, Ming Ding, Yujie Qian, Zhilin Yang, and Jie Tang. 2021{\natexlab{b}}.
\newblock Gpt understands, too.
\newblock \emph{arXiv preprint arXiv:2103.10385}.

\bibitem[{Malo et~al.(2014)Malo, Sinha, Korhonen, Wallenius, and Takala}]{Malo2014fpb}
P.~Malo, A.~Sinha, P.~Korhonen, J.~Wallenius, and P.~Takala. 2014.
\newblock Good debt or bad debt: Detecting semantic orientations in economic texts.
\newblock \emph{Journal of the Association for Information Science and Technology}, 65.

\bibitem[{Mihaylov et~al.(2018)Mihaylov, Clark, Khot, and Sabharwal}]{mihaylov2018openbookqa}
Todor Mihaylov, Peter Clark, Tushar Khot, and Ashish Sabharwal. 2018.
\newblock Can a suit of armor conduct electricity? a new dataset for open book question answering.
\newblock In \emph{Proceedings of the 2018 Conference on Empirical Methods in Natural Language Processing}, pages 2381--2391.

\bibitem[{Pal et~al.(2022)Pal, Umapathi, and Sankarasubbu}]{pal2022medmcqa}
Ankit Pal, Logesh~Kumar Umapathi, and Malaikannan Sankarasubbu. 2022.
\newblock Medmcqa: A large-scale multi-subject multi-choice dataset for medical domain question answering.
\newblock In \emph{Conference on health, inference, and learning}, pages 248--260. PMLR.

\bibitem[{Sakaguchi et~al.(2021)Sakaguchi, Bras, Bhagavatula, and Choi}]{sakaguchi2021winogrande}
Keisuke Sakaguchi, Ronan~Le Bras, Chandra Bhagavatula, and Yejin Choi. 2021.
\newblock Winogrande: An adversarial winograd schema challenge at scale.
\newblock \emph{Communications of the ACM}, 64(9):99--106.

\bibitem[{Sap et~al.(2019)Sap, Rashkin, Chen, Le~Bras, and Choi}]{sap2019socialiqa}
Maarten Sap, Hannah Rashkin, Derek Chen, Ronan Le~Bras, and Yejin Choi. 2019.
\newblock Social iqa: Commonsense reasoning about social interactions.
\newblock In \emph{Proceedings of the 2019 Conference on Empirical Methods in Natural Language Processing and the 9th International Joint Conference on Natural Language Processing (EMNLP-IJCNLP)}, pages 4463--4473.

\bibitem[{Shi and Lipani(2023)}]{shi2023dept}
Zhengxiang Shi and Aldo Lipani. 2023.
\newblock Dept: Decomposed prompt tuning for parameter-efficient fine-tuning.
\newblock \emph{arXiv preprint arXiv:2309.05173}.

\bibitem[{Si et~al.(2024)Si, Yang, and Shen}]{DBLP:journals/corr/abs-2407-05417}
Chongjie Si, Xiaokang Yang, and Wei Shen. 2024.
\newblock \href {https://doi.org/10.48550/ARXIV.2407.05417} {See further for parameter efficient fine-tuning by standing on the shoulders of decomposition}.
\newblock \emph{CoRR}, abs/2407.05417.

\bibitem[{Sun et~al.(2024)Sun, Zhuang, Wei, Zhang, and Dai}]{sun2024bbox}
Haotian Sun, Yuchen Zhuang, Wei Wei, Chao Zhang, and Bo~Dai. 2024.
\newblock Bbox-adapter: Lightweight adapting for black-box large language models.
\newblock In \emph{International Conference on Machine Learning}, pages 47280--47304. PMLR.

\bibitem[{Sun et~al.(2022{\natexlab{a}})Sun, He, Qian, Zhou, Huang, and Qiu}]{sun2022bbtv2}
Tianxiang Sun, Zhengfu He, Hong Qian, Yunhua Zhou, Xuan-Jing Huang, and Xipeng Qiu. 2022{\natexlab{a}}.
\newblock Bbtv2: Towards a gradient-free future with large language models.
\newblock In \emph{Proceedings of the 2022 Conference on Empirical Methods in Natural Language Processing}, pages 3916--3930.

\bibitem[{Sun et~al.(2022{\natexlab{b}})Sun, Shao, Qian, Huang, and Qiu}]{sun2022black-box}
Tianxiang Sun, Yunfan Shao, Hong Qian, Xuanjing Huang, and Xipeng Qiu. 2022{\natexlab{b}}.
\newblock Black-box tuning for language-model-as-a-service.
\newblock In \emph{International Conference on Machine Learning}, pages 20841--20855. PMLR.

\bibitem[{Taori et~al.(2023)Taori, Gulrajani, Zhang, Dubois, Li, Guestrin, Liang, and Hashimoto}]{alpaca}
Rohan Taori, Ishaan Gulrajani, Tianyi Zhang, Yann Dubois, Xuechen Li, Carlos Guestrin, Percy Liang, and Tatsunori~B. Hashimoto. 2023.
\newblock Stanford alpaca: An instruction-following llama model.
\newblock \url{https://github.com/tatsu-lab/stanford_alpaca}.

\bibitem[{Touvron et~al.(2023)Touvron, Martin, Stone, Albert, Almahairi, Babaei, Bashlykov, Batra, Bhargava, Bhosale et~al.}]{touvron2023llama}
Hugo Touvron, Louis Martin, Kevin Stone, Peter Albert, Amjad Almahairi, Yasmine Babaei, Nikolay Bashlykov, Soumya Batra, Prajjwal Bhargava, Shruti Bhosale, et~al. 2023.
\newblock Llama 2: Open foundation and fine-tuned chat models.
\newblock \emph{arXiv preprint arXiv:2307.09288}.

\bibitem[{Wen and Chaudhuri(2024)}]{wen2023batched}
Yeming Wen and Swarat Chaudhuri. 2024.
\newblock Batched low-rank adaptation of foundation models.
\newblock \emph{ICLR 2024}.

\bibitem[{Wu et~al.(2022)Wu, Wang, Gu, Hou, Dong, Vydiswaran, and Ma}]{wu2022idpg}
Zhuofeng Wu, Sinong Wang, Jiatao Gu, Rui Hou, Yuxiao Dong, VG~Vinod Vydiswaran, and Hao Ma. 2022.
\newblock Idpg: An instance-dependent prompt generation method.
\newblock In \emph{Proceedings of the 2022 Conference of the North American Chapter of the Association for Computational Linguistics: Human Language Technologies}, pages 5507--5521.

\bibitem[{Xiao et~al.(2023)Xiao, Lin, and Han}]{xiao2023offsite}
Guangxuan Xiao, Ji~Lin, and Song Han. 2023.
\newblock Offsite-tuning: Transfer learning without full model.
\newblock \emph{arXiv preprint arXiv:2302.04870}.

\bibitem[{Yang et~al.(2024)Yang, Yang, Zhang, Hui, Zheng, Yu, Li, Liu, Huang, Wei et~al.}]{yang2024qwen2}
An~Yang, Baosong Yang, Beichen Zhang, Binyuan Hui, Bo~Zheng, Bowen Yu, Chengyuan Li, Dayiheng Liu, Fei Huang, Haoran Wei, et~al. 2024.
\newblock Qwen2. 5 technical report.
\newblock \emph{arXiv preprint arXiv:2412.15115}.

\bibitem[{Yang et~al.(2025)Yang, Jia, Gu, Lin, Chen, Pang, Yin, Sun, Wu, and Wang}]{yang2025orthogonalfinetuningdirectpreference}
Chenxu Yang, Ruipeng Jia, Naibin Gu, Zheng Lin, Siyuan Chen, Chao Pang, Weichong Yin, Yu~Sun, Hua Wu, and Weiping Wang. 2025.
\newblock \href {https://arxiv.org/abs/2409.14836} {Orthogonal finetuning for direct preference optimization}.
\newblock \emph{Preprint}, arXiv:2409.14836.

\bibitem[{Yang et~al.(2023)Yang, Liu, and Dan~Wang}]{yang2023fingpt}
Hongyang Yang, Xiao-Yang Liu, and Christina Dan~Wang. 2023.
\newblock Fingpt: Open-source financial large language models.
\newblock \emph{FinLLM at IJCAI}.

\bibitem[{Zellers et~al.(2019)Zellers, Holtzman, Bisk, Farhadi, and Choi}]{zellers2019hellaswag}
Rowan Zellers, Ari Holtzman, Yonatan Bisk, Ali Farhadi, and Yejin Choi. 2019.
\newblock Hellaswag: Can a machine really finish your sentence?
\newblock In \emph{Proceedings of the 57th Annual Meeting of the Association for Computational Linguistics}, pages 4791--4800.

\bibitem[{Zheng et~al.(2024)Zheng, Tan, Li, and Liu}]{zheng2024bpt-subspace}
Yuanhang Zheng, Zhixing Tan, Peng Li, and Yang Liu. 2024.
\newblock Black-box prompt tuning with subspace learning.
\newblock \emph{IEEE/ACM Transactions on Audio, Speech, and Language Processing}.

\end{thebibliography}

\clearpage

\appendix

\section{Training Details}
\label{sec:appendix-a}

\subsection{Prompt Template}
\label{sec:appendix-a-prompt-template}
Our prompt template design strictly adheres to Alpaca \cite{alpaca}, uniformly applied across all datasets through the following structured template:
\begin{center}
    \fbox{
        \parbox{0.9\linewidth}{
            \textit{Below is an instruction that describes a task. Write a response that appropriately completes the request.}
            
            \noindent
            \textit{\#\#\# Instruction: \{instruction\}}
            
            \noindent
            \textit{\#\#\# Input: \{input\}}
            
            \noindent
            \textit{\#\#\# Response: }
        }
    }
\end{center}

\begin{table*}[t]
    \centering
    \resizebox{\textwidth}{!}{
    \begin{tabular}{l|l|p{8cm}|l}
        \hline
        \textbf{Domain} & \textbf{Dataset} & \textbf{Instruction} & \textbf{Input} \\
        \hline
        \multirow{1}{*}{CR} & BoolQ & Please answer the following question with true or false, question: [QUESTION] \newline Answer format: true/false & None \\
        \hline
        \multirow{3}{*}{CR} & Others & Please choose the correct answer to the question: [QUESTION] \newline Answer1: [ANSWER\_1] \newline Answer2: [ANSWER\_2] \newline Answer3: [ANSWER\_3] \newline Answer4: [ANSWER\_4] \newline Answer format: answer1/answer2/answer3/answer4 & None \\
        \hline
        Medical & All Datasets & The following are multiple choice questions about \{subject\_name\}. Output a single option from the options as the final answer. QUESTION: [QUESTION] \newline Answer1: [ANSWER\_1] \newline Answer2: [ANSWER\_2] \newline Answer3: [ANSWER\_3] \newline Answer4: [ANSWER\_4] \newline Answer format: answer1/answer2/answer3/answer4 & None \\
        \hline
        Financial & All Datasets & What is the sentiment of this news? Please choose an answer from \{negative/neutral/positive\}. & Question Text \\
        \hline
    \end{tabular}
    }
    \caption{Prompt Template Configuration for Different Domains. Note that in the Commonsense Reasoning (CR) domain, BoolQ has a distinct instruction template, while Arc-e, Arc-c, and OBQA share a common instruction template.}
    \label{tab:prompt_template}
\end{table*}

\begin{table}[t]
    \centering
    \resizebox{\linewidth}{!}{
    \begin{tabular}{llcccc}
        \toprule
        Domain & Model & Data Size & Budget $B$ & Population $\lambda$ & Step Size $\sigma$ \\
        \midrule
        \multirow{1}{*}{Commonsense} 
        & LLaMA-2-7B       & 16 & 300  & 30 & 0.01 \\
        \midrule
        \multirow{3}{*}{Medical}    
        & LLaMA-2-7B       & 5  & 1500 & 30 & 0.05 \\
        & Qwen-2.5-3B      & 5  & 300  & 30 & 0.01 \\
        & LLaMA-2-13B      & 5  & 1500 & 30 & 0.01 \\
        \midrule
        \multirow{3}{*}{Financial} 
        & LLaMA-2-7B       & 16 & 300  & 30 & 0.05 \\
        & Qwen-2.5-3B      & 16 & 300  & 30 & 0.01 \\
        & LLaMA-2-13B      & 16 & 300  & 30 & 0.05 \\
        \bottomrule
    \end{tabular}
    }
    \caption{Hyperparameter settings for the ULC stage across domains and models.}
    \label{tab:blackbox_hyperparams}
\end{table}

\begin{table}[t]
    \centering
    \resizebox{\linewidth}{!}{
    \begin{tabular}{llcccc}
        \toprule
        Domain & Setting & Learning Rate & Batch Size & Epochs & Cutoff Length \\
        \midrule
        \multirow{2}{*}{Commonsense} 
        & Prompt Tuning       & 3e-2  & 16 & 5 & 512 \\
        & CBP-Tuning          & 3e-4  & 16 & 5 & 512 \\
        \midrule
        \multirow{2}{*}{Medical}    
        & Prompt Tuning       & 3e-2  & 16 & 5 & 256 \\
        & CBP-Tuning          & 3e-4  & 16 & 5 & 256 \\
        \midrule
        \multirow{2}{*}{Financial} 
        & Prompt Tuning       & 1e-2  & 16 & 2 & 512 \\
        & CBP-Tuning          & 1e-4  & 16 & 2 & 512 \\
        \bottomrule
    \end{tabular}
    }
    \caption{Hyperparameter settings for Prompt Tuning and CBP-Tuning across different domains.}
    \label{tab:hyperparams}
\end{table}

For the setup of the instruction, in the commonsense reasoning domain, we utilize datasets processed by LLM-Adapters \cite{hu2023llm-adapters}; in the medical domain, we follow the template from Aloe \cite{gururajan2024aloe}; in the financial domain, we follow the settings of FinGPT \cite{yang2023fingpt}. The Alpaca-style prompt templates to be filled for the three domains are shown in Table \ref{tab:prompt_template}, where the \{subject\_name\} is the task name of the MMLU subset. For the commonsense reasoning and medical domains, the output format for evaluation and training is 'the correct answer is [ANSWER]' for the LLaMA-2-7B and LLaMA-2-13B, and '[ANSWER]' for the Qwen-2.5-3B. For the financial domain, the output format for evaluation and training is '[ANSWER]' for all models. It is important to note that only in the financial domain do we need to fill both the instruction and the input; in the other two domains, only the instruction is required.

\subsection{Hyperparameters}
\label{sec:appendix-a-hyperparameters}
The hyperparameters for Prompt Tuning and CBP-Tuning SDT stage in the three domains are shown in the Table \ref{tab:hyperparams}.

The hyperparameters for CBP-Tuning ULC stage in the three domains are shown in the Table \ref{tab:blackbox_hyperparams}.

\subsection{Hardware}
\label{sec:appendix-a-hardware}
We use 2 NVIDIA A100 40G GPUs to conduct our experiments, and the version of transformers library is 4.41.2, while the version of PEFT is 0.3.0. For SDT in the commonsense reasoning domain with LLaMA-2-7B, approximately 12 GPU hours are required.

\subsection{Environment of Experiments}
\label{sec:appendix-a-env}
The implementation of CBP-Tuning is based on Transformers library, peft library and LLM-Adapters. The data processing and evaluation for two domains also follow LLM-Adapters. 

\begin{table*}[t]
    \centering
    \resizebox{\textwidth}{!}{
    \begin{tabular}{lccccccc}
        \toprule
        Method & CK & MG & Anatomy & PM & CB & CM & Avg. \\
        \midrule
        CBP-Tuning (dim=100) w/o ULC 
            & 53.21 & 52.00 & 54.07 & 58.82 & 67.36 & 49.13 & 55.77 \\
        CBP-Tuning (dim=100, $\sigma$=0.01, $B$=300) 
            & 57.86 & 56.50 & 58.52 & 60.05 & 69.44 & 51.45 & 58.97 \\
        \midrule
        CBP-Tuning (dim=1000) w/o ULC 
            &  4.15 & 17.00 &  4.44 &  9.19 &  1.39 &  6.36 &  7.09 \\
        CBP-Tuning (dim=1000, $\sigma$=0.05, $B$=300) 
            &  4.03 & 15.33 &  3.46 & 26.72 & 21.30 & 22.74 & 15.59 \\
        \midrule
        CBP-Tuning (dim=500) w/o ULC (ours) 
            & 74.00 & 59.26 & 66.54 & 74.31 & 61.85 & 67.31 & 67.92 \\
        CBP-Tuning (dim=500, $\sigma$=0.01, $B$=300) (ours) 
            & 74.00 & 59.26 & 69.49 & 74.31 & 63.39 & 68.17 & 68.55 \\
        \bottomrule
    \end{tabular}
    }
    \caption{Ablation on bottleneck dimension $r$ for Qwen-2.5-3B in the medical domain.}
    \label{tab:dim_ablation}
\end{table*}

\section{Additional Results}
\label{sec:appendix-b}

\subsection{Decreased Performance Analysis}
\label{sec:appendix-b-analysis}

The table \ref{tab:cm_optimization_losses} shows the final loss values to which each dataset converged during the user-side local customization (ULC) stage using the CMA-ES algorithm (lower values indicate more complete optimization). Notably, the final loss for MMLU Clinical Medicine (CM) is $1.361\times10^{-2}$, three orders of magnitude higher than that for Medical Genetics (MG), which is $2.972\times10^{-5}$. This indicates that the optimization process for the CM task was much harder to converge, consistent with the performance drop we observed for this task in the main text.

\begin{table*}[t]
    \centering
    \resizebox{\textwidth}{!}{
    \begin{tabular}{lcccccc}
        \toprule
        & CK & MG & Anatomy & PM & CB & CM \\
        \midrule
        Loss 
        & $2.078\times10^{-4}$ 
        & $2.972\times10^{-5}$ 
        & $3.496\times10^{-4}$ 
        & $5.906\times10^{-3}$ 
        & $2.873\times10^{-3}$ 
        & $1.361\times10^{-2}$ \\
        \bottomrule
    \end{tabular}
    }
    \caption{Final CMA-ES loss values at the end of the ULC stage for each medical dataset using LLaMA-2-7B.}
    \label{tab:cm_optimization_losses}
\end{table*}

\begin{table*}[t]
    \centering
    \resizebox{\textwidth}{!}{
    \begin{tabular}{lcccccccclcccc}
        \toprule
        \multirow{2}{*}{Model} & \multirow{2}{*}{Setting} 
        & \multicolumn{7}{c}{Medical Domain}
        & & \multicolumn{4}{c}{Financial Domain} \\
        \cmidrule{3-9} \cmidrule{11-14}
        & & CK & MG & Anatomy & PM & CB & CM & Med Avg & & FIQA\_SA & TFNS & FPB & Fin Avg \\
        \hline
        \multirow{6}{*}{\textbf{Qwen-2.5-3B}} 
        & Zero Shot & 39.25 & 49.00 & 49.63 & 20.59 & 36.81 & 39.31 & 39.10 & & 66.67 & 38.65 & 77.96 & 61.09 \\
        & Prompt Tuning & 65.66 & 71.00 & 54.81 & 37.87 & 67.36 & 60.12 & 59.47 & & 58.12 & 69.26 & \textbf{83.97} & 70.45 \\
        & CBP-Tuning w/o ULC & 67.92 & \textbf{74.00} & 59.26 & 66.54 & \textbf{74.31} & 61.85 & 67.31 & & 62.39 & 69.85 & 72.44 & 68.53 \\
        & \textbf{CBP-Tuning} & \textbf{68.55} & \textbf{74.00} & 59.26 & \textbf{69.49} & \textbf{74.31} & \textbf{63.39} & \textbf{68.17} & & 72.22 & 70.45 & 77.80 & \textbf{73.49} \\
        \cmidrule{2-14}
        & LoRA & 64.15 & 73.00 & 58.52 & 59.93 & 66.67 & 52.60 & 62.48 & & \textbf{79.49} & \textbf{72.91} & 64.62 & 72.34 \\
        & P-Tuning & 69.06 & 71.00 & \textbf{60.00} & 57.35 & 72.92 & 61.85 & 65.36 & & 24.36 & 69.72 & 75.93 & 56.67 \\
        \bottomrule
    \end{tabular}
    }
    \caption{Performance (\%) of Qwen-2.5-3B compared with additional baselines (LoRA and P-Tuning) across medical and financial domains. \textbf{Bold} indicates the best performance in each column among all methods.}
    \label{tab:appendix_baselines}
\end{table*}

\subsection{Bottleneck Dimension Ablation Study}
\label{sec:appendix-b-dim}

In Table \ref{tab:dim_ablation} we compare three bottleneck dimensions ($r=100,500,1000$) both without and with the ULC phase using LLaMA-2-7B in the medical domain. The results peak at $r=500$, where CBP-Tuning achieves an average accuracy of 68.55\%, significantly higher than the 58.97\% at $r=100$ and only 15.59\% at $r=1000$. This confirms that 500 provides the best trade-off between representational capacity and optimization stability.

\subsection{Performance Comparison with LoRA and P-Tuning}
\label{sec:appendix-b-baselines}

We conducted additional experiments comparing CBP-Tuning with other prominent parameter-efficient fine-tuning (PEFT) methods: LoRA \cite{hu2021lora} and P-Tuning \cite{liu2021p-tuning}. The comparison was performed on the Qwen-2.5-3B model across the medical and financial domains.

As shown in the table \ref{tab:appendix_baselines}, CBP-Tuning demonstrates superior average performance compared to both LoRA and P-Tuning. In the medical domain, CBP-Tuning achieves performance improvements of 5.69 points and 2.81 points over LoRA and P-Tuning in average scores, respectively. In the financial domain, the improvements are 1.15 points and 16.82 points, respectively, further validating the effectiveness of our approach.

Furthermore, it is important to note the architectural differences. CBP-Tuning is designed based on Prompt Tuning, as both methods modify only the token embeddings at the input layer. In contrast, LoRA requires adapting and merging low-rank matrices within each layer of the model. This makes the comparison between CBP-Tuning and Prompt Tuning particularly direct, while highlighting CBP-Tuning’s potential for higher efficiency in deployment scenarios where modifying internal model weights is more complex or costly.

\end{document}